\documentclass[letterpaper, 10 pt, conference]{ieeeconf}  

\IEEEoverridecommandlockouts                              

\usepackage[ruled, noend, linesnumbered]{algorithm2e}
\usepackage{amsmath}
\usepackage{amssymb}
\usepackage{dsfont}
\usepackage{bm}
\usepackage{tabularx, booktabs}
\usepackage{multicol}
\usepackage{multirow}
\usepackage{dblfloatfix} 
\usepackage{graphicx}
\usepackage{xcolor}
\usepackage[footnotesize]{caption}

\usepackage{subfig}

\newcommand{\dR}{d_{\mathcal{R}}}
\DeclareMathOperator*{\argmin}{argmin}

\renewcommand{\vec}[1]{\bm{#1}}

\title{\LARGE \bf
Stochastic Modeling of Distance to Collision for Robot Manipulators
}

\author{Nikhil Das$^{1}$ and Michael C. Yip$^{1}$
\thanks{$^{1}$Nikhil Das and Michael C. Yip are with the Department of Electrical and Computer Engineering, University of California San Diego
        {\tt\small \{nrdas,yip\}@ucsd.edu}}%
}

\begin{document}

\maketitle
\thispagestyle{empty}
\pagestyle{empty}


\begin{abstract}
Evaluating distance to collision for robot manipulators is useful for assessing the feasibility of a robot configuration or for defining safe robot motion in unpredictable environments. However, distance estimation is a time-consuming operation, and the sensors involved in measuring the distance are always noisy. A challenge thus exists in evaluating the expected distance to collision for safer robot control and planning. In this work, we propose the use of Gaussian process (GP) regression and the forward kinematics (FK) kernel (a similarity function for robot manipulators) to efficiently and accurately estimate distance to collision. We show that the GP model with the FK kernel achieves 70 times faster distance evaluations compared to a standard geometric technique, and up to 13 times more accurate evaluations compared to other regression models, even when the GP is trained on noisy distance measurements. We employ this technique in trajectory optimization tasks and observe 9 times faster optimization than with the noise-free geometric approach yet obtain similar optimized motion plans. We also propose a confidence-based hybrid model that uses model-based predictions in regions of high confidence and switches to a more expensive sensor-based approach in other areas, and we demonstrate the usefulness of this hybrid model in an application involving reaching into a narrow passage.
\end{abstract}

\section{Introduction}
Evaluating whether a given configuration of a robot manipulator is in a feasible position, such as for motion planning applications, typically involves a collision detector that produces a binary output: in-collision or collision-free \cite{Das2020}. However, it is often far more useful to know the distance to collision in the workspace, as closer proximities may suggest that the robot is more at risk to colliding with an obstacle. This distance to collision is useful in defining robot control rules for safe robot motion in potentially unpredictable, partially observed, or nonstationary environments \cite{Flacco2012}, such as in cases where the robots are working in close proximity with humans \cite{Schiavi2009}. In addition to collision avoidance for robots \cite{Flacco2012}, applications requiring accurate representation of the distance between two objects include contact resolution in simulation (where objects must respond to collisions realistically) \cite{Pan2013} and force feedback in haptic rendering (where forces are generated for a human manipulating a virtual or teleoperated tool) \cite{Lin2003}.

\begin{figure}[t]
	\centering
    \subfloat[Workspace\label{fig:distWorkspace}] {
		\includegraphics[width=0.46\linewidth, trim={3cm 2cm 2.2cm 1.5cm}, clip]{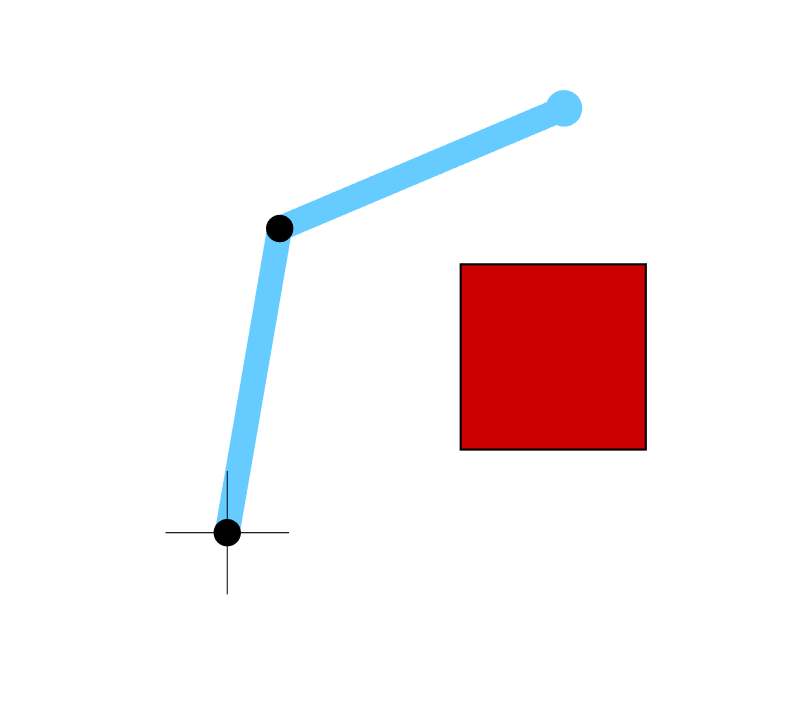}}
		\hfill
    \subfloat[Distances in C-space\label{fig:distCspace}] {
        \includegraphics[width=0.51\linewidth, trim={0cm 0cm 0cm 0cm}, clip]{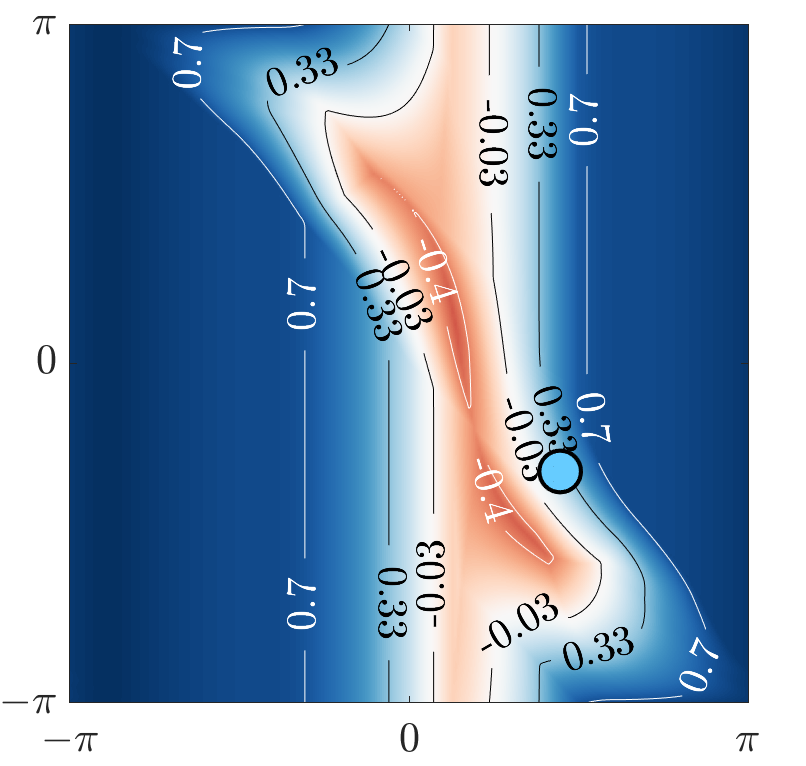}}
    \caption{(a) A 2 DOF robot and a square obstacle. (b) A C-space representation where the workspace distance to collision for each configuration is shown. The axes represent each DOF of the robot.}
    \label{fig:distExample}
\end{figure}

A robot's distance to collision is typically defined to be the minimum distance between all of the robot's links and all of the workspace obstacles \cite{Flacco2012,Han2019}. Assuming that the $i^{th}$ link of a robot in configuration $\vec{x}$ is represented as ${\mathcal{R}}_i(\vec{x})$ and the $j^{th}$ workspace obstacle is $o_j$, we define $\dR(\cdot)$ as
\begin{equation}
\dR(\vec{x}) = \min_{i,j}dist({\mathcal{R}}_i(\vec{x}), o_j)
\end{equation}
where $dist(\cdot, \cdot)$ provides a positive scalar distance between two non-intersecting bodies. When the two bodies are in collision, $dist(\cdot, \cdot)$ provides the penetration depth (the minimum amount intersecting bodies must move to be out of collision \cite{Lin2003}) as a negative value. Fig. \ref{fig:distWorkspace} shows a 2 degrees-of-freedom (DOF) robot with an obstacle, and its corresponding configuration space (or C-space) representation in Fig. \ref{fig:distCspace}. For each point in the C-space, the workspace distance to collision is shown.

As with collision detection, distance estimation is a time-consuming operation, with time complexity as high as $\mathcal{O}(m^6n^6)$ for generalized distance calculations, where $m$ and $n$ are the number of vertices in two different bodies \cite{Pan2012}. Furthermore, reliable collision avoidance often requires the extensive use of external sensors \cite{Flacco2012}. The sensors that provide obstacle information, such as cameras or depth sensors, are a source of uncertainty in the distance measurements \cite{Kyriakopoulos1990}, and thus allow us to acquire only noisy evaluations of distance, which we represent as $\tilde{d}_\mathcal{R}(\cdot)$. Accounting for this uncertainty is a necessity in distance estimation, yet few strategies exist that consider these imperfections in distance sensing. As there are many applications that utilize distances, in this work, we seek to create a model to efficiently and accurately estimate $d_{\mathcal{R}}(\cdot)$ while accounting for potential uncertainty in the distance measurements.

\subsection{Contributions}
In this work, we describe a modeling approach to estimate a robot manipulator's distance to collision given noisy measurements. More specifically, we employ Gaussian process (GP) regression with the forward kinematics (FK) kernel, a similarity function designed for robot manipulators, to estimate $\dR(\vec{x})$.

We demonstrate that the GP model with the FK kernel has significantly better modeling accuracy than other comparable regression techniques and show that querying the proposed GP is orders of magnitude faster than the standard geometric approach for distance evaluation. We utilize our proposed technique in trajectory optimization applications and realize a 9 times improvement in optimization time compared to a noise-free geometric approach yet with similar final results. Finally, we show how a hybrid model, which considers the confidence of the model, can switch from GP-based predictions to sensor-based distance estimation in areas of low confidence.

\subsection{Related Work}
The standard approach to calculating distances involves first using sensors to obtain the locations of obstacles, and then approximating the robot geometry and workspace obstacles to actually compute the distance, which is a considerably expensive process if there are many pairwise checks to perform \cite{Han2019,Safeea2017}. Some techniques attempt to reduce this computational effort through approximate methods, such as by using values from a depth sensor  to approximate a distance or by creating a proxy machine learning model using exemplar data. This section describes some geometric, depth perception-based, and machine learning techniques.


\subsubsection{Geometric Approaches}
A standard approach to calculating distance to collision is to approximate the objects and use a geometric distance method between them. For example, Han et al. \cite{Han2019} estimate $d_\mathcal{R}(\cdot)$ by first representing each link of a robot arm as a cylinder and converting point clouds acquired from a depth camera into multiple small cubes, and then using the GJK algorithm \cite{Gilbert1988} to compute the distances. While their technique can accurately represent distances for a robot and obstacles observed with a depth camera, breaking obstacles into numerous small polyhedra may increase the potential number of pairwise checks that need to be performed, and there is no measure of uncertainty with this approach.


\subsubsection{Perception-Based Distance Fields}
Flacco et al. create a method to estimate distance to collision for robot manipulators based on depth sensors \cite{Flacco2012}. For any point in the workspace, distance equations are defined using a depth map, which is an image with depth values evaluated for each pixel.

The distances are used to create artificial repulsive forces to drive the robot away from obstacles. The authors claim that the distance values computed with their technique are satisfactory for the purpose of collision avoidance but are not to be considered as actual Cartesian distances. This limitation precludes the usage of this technique beyond the purpose of defining repulsive forces.

Furthermore, their distance estimation technique has no measure of sensor or model noise. However, as the primary purpose of their technique is to define repulsive forces away from obstacles, the authors take an average over all obstacles' repulsive fields to reduce any potential noise originating from the depth sensor.


\subsubsection{Support Vector Machines}
Pan et al. use a support vector machine (SVM) to model the collision region in C-space between two bodies. Binary collision checks are performed on random configurations of one object relative to the other to form a decision boundary. By determining the distance between a query configuration and the decision boundary, their SVM model is used to estimate distance. Since there is no closed-form solution for the distance to the boundary for a nonlinear SVM, sequential quadratic programming, a constrained nonlinear optimization technique, is used to find this distance \cite{Pan2013}. As distances in the C-space do not generally equate to distances in the workspace, a different distance metric is used (the displacement distance metric), which involves determining application-specific weighting terms \cite{Pan2013}. Additionally, this approach requires a unique SVM model between each pair of obstacles, which may make applying this technique to robot manipulators or environments with multiple obstacles more computationally expensive. Finally, this technique assumes the state of each object is perfectly known, which means there is no measure of uncertainty in this model.

\subsubsection{Neural Networks}
RelaxedIK estimates self-collision costs by using line segments to represent the links. Distances between the links are used to approximate the distances. These distance computations are further approximated using a neural network \cite{Rakita2018}. Their method successfully estimates these self-collision costs which enables usage in optimization problems. However, as with many neural network approaches, this technique requires a large amount of labeled data and a significant amount of time to train, precluding this approach from being applied to instances beyond distance to self-collision.

ClearanceNet is a promising neural network-based approach to estimate $d_{\mathcal{R}}(\cdot)$. The input to this network is the robot configuration and a parameterized representation of the environment. Their approach achieves over $90\%$ accuracy if the network's output is used to predict the collision status \cite{Kew2019}. Neural networks, however, are of fixed size and complexity, often require a large amount of data to train, and since the locations of the workspace obstacles are parameterized, generalizing the implementation to new types of environments may require a restructuring of this specific neural network implementation.

\section{Background}
\subsection{Kernel Functions}
A kernel function $k:\mathbb{R}^D \times \mathbb{R}^D\rightarrow \mathbb{R}$ is a function that produces a similarity score between two objects, such as $D$-dimensional robot configurations. A kernel function is a fundamental tool for learning-based algorithms including binary classification, probability density estimation, and dimensionality reduction \cite{Hofmann2008}. Kernel functions are often employed in nonparametric models, which are models that make no strong assumptions on the structure of the model and whose complexity scales with the data \cite{Altman1992}, which is desired for modeling distance fields given that many of the environments in which a manipulator is used is nonstationary.

A kernel matrix $\vec{K}\in\mathbb{R}^{N\times N}$ can be defined for a set of $N$ points $\mathcal{X}$ where $\vec{K}_{ij} = k(\mathcal{X}_i,\mathcal{X}_j)$. A positive definite kernel is one that yields a positive definite $\vec{K}$. Positive definiteness is an important property because it guarantees nonsingularity for the kernel matrix, represents an implicit mapping to some higher-dimensional feature space according to Mercer's theorem \cite{Minh}, and is a requirement in some machine learning algorithms including Gaussian processes \cite{Rasmussen2006}.

There are various positive definite kernel functions, and the one employed in the model should be selected to capture meaningful similarities between the data \cite{BelancheMunoz2013}. Two kernels that we will use in this paper are the Gaussian kernel and the forward kinematics kernel, which are described below.

\begin{figure}[!t]
	\centering
	\subfloat[Gaussian Kernel\label{fig:gaussianKernel}]{%
       \includegraphics[width=0.32\linewidth, trim={0 0 0 0}, clip]{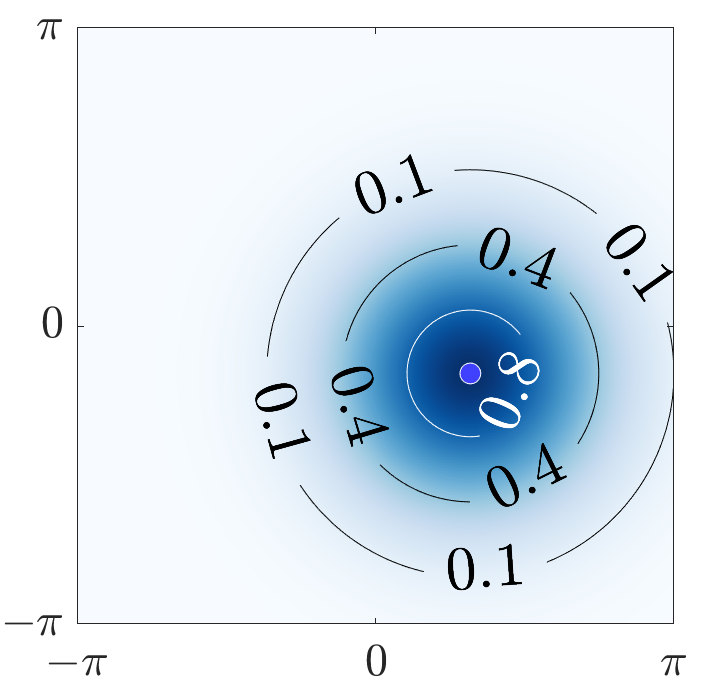}}
    \hfill
    \subfloat[FK Kernel\label{fig:fkKernel}]{%
        \includegraphics[width=0.32\linewidth, trim={0 0 0 0}, clip]{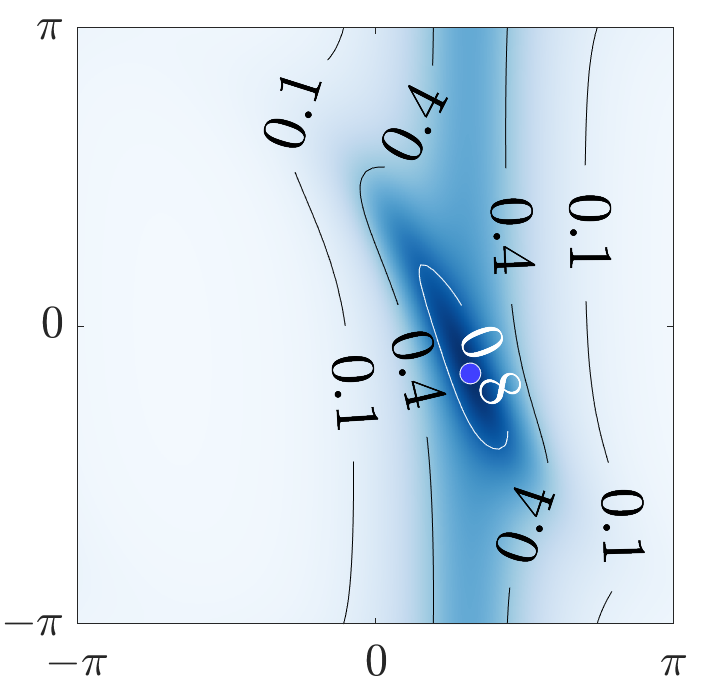}}
	\hfill
    \subfloat[2 DOF Robot\label{fig:controlPointPlacement}]{%
       {\includegraphics[width=0.32\linewidth, trim={1cm 1.1cm 1.2cm 0.2cm}, clip]{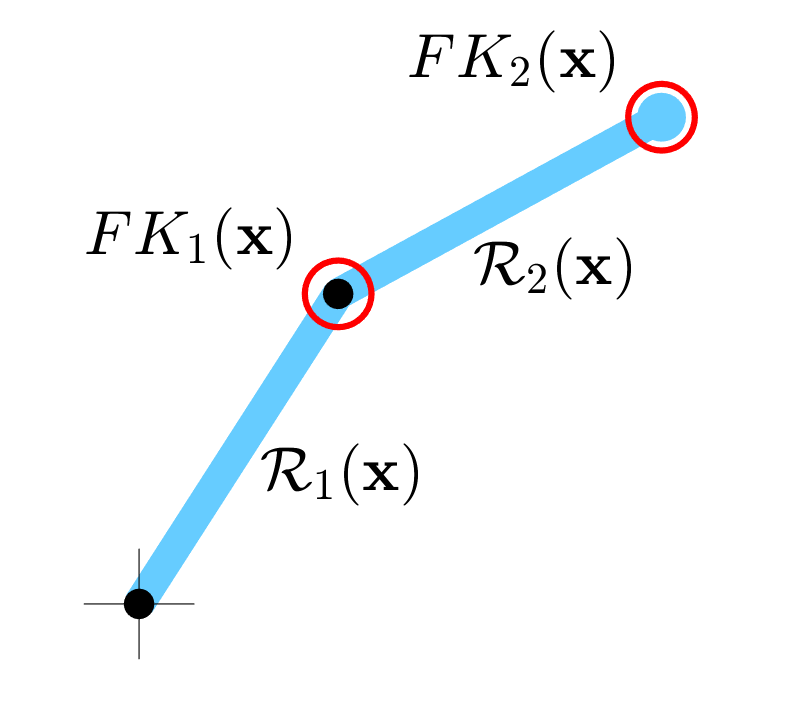}}}
    \caption[Different levels of confidence of the GP models for $d_{\mathcal{R}}(\cdot)$ being greater than 0 for a 2 DOF robot are shown.]{Visualizations of (a) the Gaussian kernel and (b) the FK kernel for the configuration of the 2 DOF robot specified in (c). The FK kernel is based on the control point locations marked with red circles.}
    \label{fig:kernelShapes}
\end{figure}

\subsubsection{Gaussian Kernel}
A common positive definite kernel function to use in kernel-based models is the Gaussian kernel $k_G(\vec{x},\vec{x'}) = \exp\left(-\gamma\|\vec{x} - \vec{x'}\|^2\right)$, where $\gamma$ is a parameter dictating the width of the kernel \cite{Rasmussen2006}. Fig. \ref{fig:gaussianKernel} shows a topological visualization of the Gaussian kernel for the 2 DOF robot in Fig. \ref{fig:controlPointPlacement}. The Gaussian kernel is often considered the default kernel to use if there is limited prior knowledge of the underlying structure of the data \cite{Smola1998}.

\subsubsection{Forward Kinematics Kernel}
The forward kinematics (FK) kernel was designed to compare robot manipulators \cite{Das2020a}. Rather than comparing the joint angles representing the robot configuration directly, the locations of $M$ control points along the arm are compared. This avoids measuring distances directly in joint space, where proximal joint displacement has far greater effect than distal joint displacement to the Cartesian shape of the robot. The FK kernel is defined as $k_{FK}(\vec{x}, \vec{x'}) = \sum_{m=1}^{M}{\left(1+\frac{\gamma}{2}\|FK_m(\vec{x}) - FK_m(\vec{x'})\|^2\right)}^{-2}$, where $FK_m(\vec{x})$ is the location of the $m^{th}$ control point in the workspace for configuration $\vec{x}$ and $\gamma$ is a parameter dictating kernel width. Fig. \ref{fig:fkKernel} shows a topological visualization of the FK kernel for a 2 DOF robot with control points placed at the end of each link. Significant improvement to classification performance was realized with the FK kernel \cite{Das2020a} over other kernels \cite{Das2020,Das2017}, and thus the FK kernel is a worthwhile avenue to explore when creating a model for distance estimation.


\subsection{Regression}
Regression models can produce a continuous scalar output given a possibly multi-dimensional input. A common perspective of regression analysis is to view the true output as a sum of its input-dependent expected value and some error term, and using the expected value as a predictor of the output for new queries \cite{Fahrmeir2013}. In the context of distance evaluation, we can assume that the measured, sensor-based distance is a sum of the true distance and a noise term: $\tilde{d}_\mathcal{R}(\vec{x}) = \dR(\vec{x}) + \epsilon$, where $\epsilon \sim \mathcal{N}(0,\eta^2)$ and $\eta^2$ is the measurement noise variance.

Choosing nonparametric models such as kernel regression and Gaussian processes allow for distance maps to scale to the complexity of the environment and to account for changing environments over time. This is the ideal scenario as many environments a robot must work in are nonstationary and may include other agents.

\subsubsection{Kernel Regression}
Kernel regression (KR) \cite{Nadaraya1964,Watson1964} is one example of a nonparametric regression technique, where a probability distribution is estimated and used to compute an expected value. A training set of $N$ input samples $\mathcal{X}$ and their corresponding outputs $\vec{y}$ can be used to estimate a probability distribution of the outputs conditioned on the inputs. This probability distribution may then be used to calculate the expected value of query inputs by taking a weighted sum of all training datapoints to predict the output of a query point. The equation for prediction for a single query point $\vec{x}$ is
\begin{equation}
\mu_{KR}\left(\vec{x}\right) = \frac{\sum_{i=1}^Nk(\mathcal{X}_i,\vec{x})\vec{y}_i}{\sum_{i=1}^Nk(\mathcal{X}_i,\vec{x})} \label{eq:kr}
\end{equation}
where $k(\cdot, \cdot)$ is some kernel function. KR does not need to learn any weights to estimate the expected value of $\dR(\cdot)$, so a training procedure is not required. KR typically uses a kernel that has equal influence in each direction, such as the Gaussian kernel \cite{Rasmussen2006}. The FK kernel is not a function of the distance between its inputs and changes shape depending on where the kernel is evaluated, so it is not well suited for the KR formulation in Eq. \ref{eq:kr}.

\subsubsection{Gaussian Process Regression}
\label{sec:gp}
Gaussian processes (GPs) can be used for regression problems. GPs are nonparametric kernel-based models that fit a Gaussian distribution over the functions that can potentially fit a set of training data \cite{Rasmussen2006}.

A GP is completely defined by a mean and covariance (or kernel) function. The prior distribution considers the space of all possible functions that can be defined using a prior mean (which is sometimes, but not always, 0) and a positive definite kernel function $k(\cdot, \cdot)$. A posterior distribution is defined given training data to condition the prior distribution. Assuming that $\mathcal{X}$ is a training dataset, $\vec{y}$ is a vector of corresponding training outputs, the posterior distribution for the regression output $y$ for a query $\vec{x}$ is
\begin{equation}
\begin{split}
P&\left(y|\mathcal{X},\vec{y},\vec{x}\right) = \mathcal{N}\left(k(\vec{x}, \mathcal{X})[k(\mathcal{X}, \mathcal{X}) +\eta^2\vec{I}]^{-1}\vec{y}, \right.\\
&\left. k(\vec{x}, \vec{x}) - k(\vec{x}, \mathcal{X})[k(\mathcal{X}, \mathcal{X}) + \eta^2\vec{I}]^{-1}k(\mathcal{X},\vec{x})\right)
\end{split} \label{eq:gpEquation}
\end{equation}
where $\eta^2$ is the measurement noise variance in the measurements of $\vec{y}$ \cite{Rasmussen2006}. This noise term can either be selected prior to model fitting if it is known or estimated in advance, or it can be treated as a hyperparameter that can be estimated based on the training dataset \cite{Rasmussen2006}.

For a single query $\vec x$, the posterior mean according to Eq. \ref{eq:gpEquation}, $\mu(\vec x)$, can serve as the regression output. A benefit of GPs is some measure of confidence may be obtained with the variance of the posterior distribution in Eq. \ref{eq:gpEquation}, represented now as $\sigma^2(\vec x)$. A confidence intervals (CI) is a set of bounds containing the true output with some probability. The bounds of a two-sided CI are $\mu(\vec x) \pm z\,\sigma(\vec x)$, where $z$ corresponds to a number of standard deviations from the mean. $z$ can be selected to contain a certain proportion of the area under a standard Gaussian distribution and may be determined from Gaussian distribution tables \cite{Altman1988} or statistics software. For example, if $z=1.96$, the two-sided CI contains 95\% of the area, which intuitively means the CI has a probability of 95\% of containing the true regression output \cite{Altman1988}. Selecting larger $z$ values to increase the probability of containing the true output will have a tradeoff of looser bounds on the CI.


The inversion that occurs in Eq. \ref{eq:gpEquation} may be costly for larger training datasets. This issue has been addressed by reduced rank approximations or using subsets of the training dataset \cite{Rasmussen2006}, which are handled by many GP software libraries, and for which robot-specific GP modeling strategies have been proposed \cite{Wilcox2020}.

\section{Methods}
In this section, we define a model $f : \mathbb{R}^D\rightarrow \mathbb{R}$ that accurately estimates $\dR(\vec{x})$ for a given robot configuration $\vec{x}$ and accounts for potential uncertainty in the estimation.

\subsection{Gaussian Process Regression Model Fitting}
We propose to fit a GP model to a training set of configurations $\mathcal{X}$ because of a GP's flexibility in modeling arbitrary datasets and its capability in accounting for uncertainty. We further propose the use of the FK kernel in the GP due to the boost in performance this kernel has proven to provide in binary proxy collision checking applications \cite{Das2020a}.

Uniformly randomly sampled robot configurations are used as the training set $\mathcal{X}$. Each configuration in the set is labeled with a noisy distance measurement (calculated using the GJK/EPA \cite{Gilbert1988,VandenBergen2001} geometric methods in our implementation), and the labels are stored in vector $\vec{y}$. To simulate noisy, sensor-based measurements, each label in $\vec{y}$ is created by taking a sum of the true distance and Gaussian noise: $\tilde{d}_\mathcal{R}(\vec{x}) = \dR(\vec{x}) + \epsilon$, where $\epsilon \sim \mathcal{N}(0,\eta^2)$.

All methods were written in MATLAB. In the implementation of the FK kernel, a control point is placed at the distal end of each link in the manipulator as was done in \cite{Das2020a}.



\subsection{Confidence-Based Hybrid Model}
\label{sec:hybrid}
As described in Section \ref{sec:gp}, two-sided CIs may be defined based on the posterior distribution standard deviation. Similarly, a one-sided CI may be defined, e.g., $[\mu(\vec x) - z\,\sigma(\vec x), \infty)$ is the CI with a lower bound on what the true regression output is for configuration $\vec{x}$ with some degree of confidence. In this case, $z=1.64$ means that there is 95\% probability that the true regression output is greater than $\mu(\vec x) - 1.64\,\sigma(\vec x)$. A one-sided lower-bounded CI may be used when a bound tighter than in the two-sided CI case is desired and an upper bound on the true value is not as important as a lower bound.


\begin{figure}[!t]
	\centering
	\subfloat[GP (Gaussian)\label{fig:ciRBF}]{%
       \includegraphics[width=0.49\linewidth, trim={0 0 0 0}, clip]{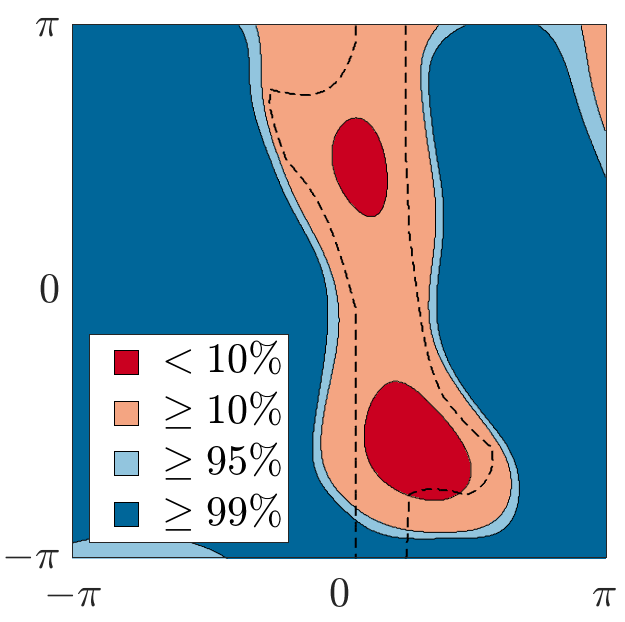}}
    \hfill
    \subfloat[GP (FK)\label{fig:ciFK}]{%
        \includegraphics[width=0.49\linewidth, trim={0 0 0 0}, clip]{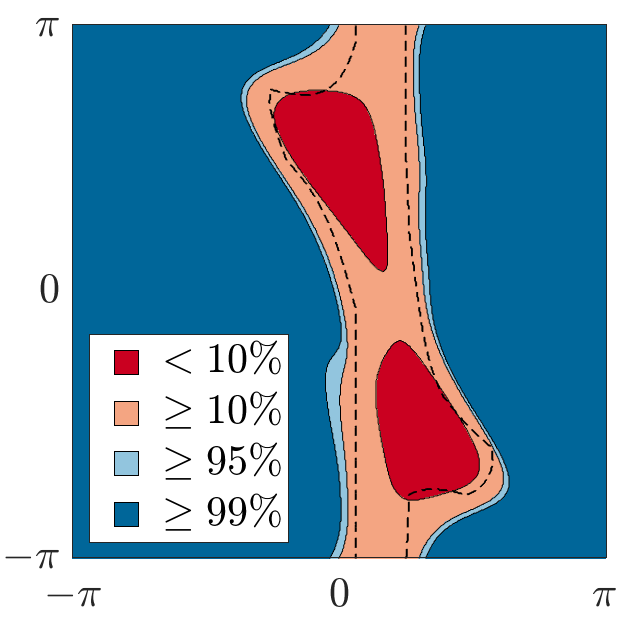}}
	    \caption[Different levels of confidence of the GP models for $d_{\mathcal{R}}(\cdot)$ being greater than 0 for a 2 DOF robot are shown.]{The colored regions represent different levels of confidence of the GP models for $d_{\mathcal{R}}(\cdot)$ being greater than 0 for a 2 DOF robot, where each axis represents one DOF. The ground truth is shown in Fig. \ref{fig:distExample}. The dashed black curve is where $d_{\mathcal{R}}(\cdot)=0$ using GJK. The GPs use (a) the Gaussian kernel and (b) the FK kernel.}
    \label{fig:ci}
\end{figure}

In Fig. \ref{fig:ci}, we visualize the levels of confidence for a 2 DOF robot having $d_{\mathcal{R}}(\vec{x})\ge 0$ when using either GP (Gaussian) or GP (FK). The dashed black curve is where $d_{\mathcal{R}}(\vec{x}) = 0$. We can see that the confidence regions for GP (FK) closely align with the C-space obstacle contour, while GP (Gaussian) does not match as closely. In fact, GP (Gaussian) has pockets of inappropriate confidence, such as decreased confidence in the corners of the shown C-space or high confidence at the bottom of the C-space obstacle. On the other hand, GP (FK) does not have these issues and has increasing confidence of $d_{\mathcal{R}}(\vec{x})$ truly being greater than 0 for configurations farther away from the C-space obstacle. GP (FK)'s more appropriate assignment of confidence suggests that we can create a hybrid model that switches distance estimation methods depending on the GP's level of confidence.


Using the lower-bounded CI, we define a hybrid distance estimation technique that uses the GP mean as the predicted distance in areas with a high confidence of being collision-free and a more expensive method in other areas:
\begin{equation}
\mu^{(P)}(\vec{x}) = 
\begin{cases}
\mu(\vec{x}), & \text{if }\mu(\vec{x}) - z\,\sigma(\vec{x}) \ge 0 \\
\frac{1}{N_{\tilde{d}}}\sum_{i=1}^{N_{\tilde{d}}}\tilde{d}_\mathcal{R}(\vec{x}), & \text{otherwise}
\end{cases} \label{eq:hybrid}
\end{equation} 
where $P$ denotes the desired level of confidence associated with the number of standard deviations $z$ (e.g., $P=95\%$ for $z=1.64$), and $N_{\tilde{d}}$ is the number of samples to collect from $\tilde{d}_\mathcal{R}(\cdot)$ to potentially reduce the effects of sensor noise. The lower bound in Eq. \ref{eq:hybrid} may of course be nonzero if a more conservative model is desired. This hybrid model may be useful when cheaper distance evaluations from the GP may be used when working far from the obstacle, but more meticulous methods may be required when in close proximity with obstacles.

\section{Results}
We evaluate the performance of our proposed GP-based distance estimation technique against three other approaches: the geometric approach GJK/EPA (ground truth), a GP with a typical Gaussian kernel, and KR. 

We begin by comparing the accuracy and timing of each technique before applying these techniques to path optimization. In each experiment, a 7 DOF planar robot manipulator is modeled with unit length link and a rectangular link profile, and a polygonal workspace obstacle is randomly placed in the manipulator's reachable workspace. We set the measurement noise to be $\eta = 0.05$ when creating each dataset. To get an idea of the level of this noise, for an obstacle one link length away from the robot (i.e., $\dR(\vec{x}) = 1$), 99\% of the measured values could be anywhere in the range $[0.85,1.15]$. We assume the scale of the robot and environments are arbitrary, so units for any measurements of length are omitted.


\subsection{Accuracy and Query Timing}
In this section, we evaluate the GP model's performance in terms of its accuracy and computation timing. Metrics that we utilize to evaluate performance include mean squared error (MSE) and model query time. MSE is defined as
\begin{equation}
MSE = \textstyle \frac{1}{\widehat{N}}\sum_{i=1}^{\widehat{N}}{\left|\widehat{\vec{y}}_i-\mu\left(\widehat{\mathcal{X}}_i\right)\right|^2} 
\label{eq:mse}
\end{equation}
where the test set $\widehat{\mathcal{X}}$ has $\widehat{N}$ elements, $\widehat{\vec{y}}$ is the set of true labels for $\widehat{\mathcal{X}}$, and $\mu(\cdot)$ provides the predicted distance using one of the approximation techniques. Note that while each model is trained on noisy distance evaluations (i.e., $\tilde{d}_\mathcal{R}(\cdot)$), we use noise-free distance evaluations (i.e., $\dR(\cdot)$) to obtain $\widehat{\vec{y}}$ when computing the MSE.

\begin{figure}[!t]
	\centering
    \includegraphics[width=\linewidth, trim={.5cm 0cm 1.4cm 0.1cm}, clip]{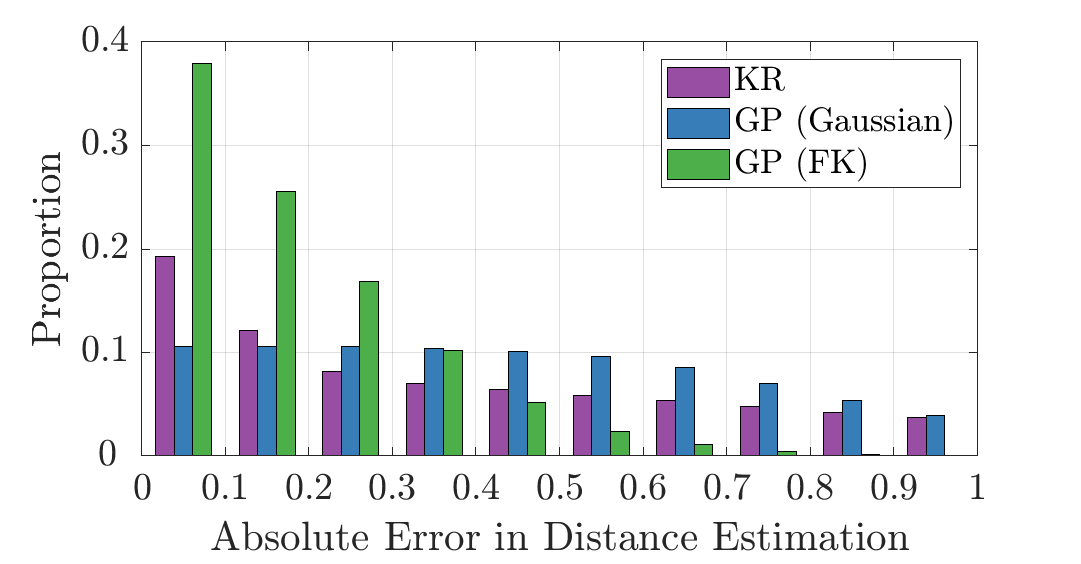}
    \caption{Histogram of absolute errors in distance estimation of $d_{\mathcal{R}}(\cdot)$ when using KR and GP models with the Gaussian and FK kernel. Models with a higher proportion of lower error are better.}
    \label{fig:errorHist}
\end{figure}

Because the volumes of the in-collision and collision-free regions in the C-space may not be equal, the overall MSE may misrepresent the MSE within each region. Thus, we also evaluate MSE on subsets of the test data for which $\widehat{\vec{y}}$ is either positive (collision-free) or negative (in-collision), which we refer to as true positive MSE (TPMSE) and true negative MSE (TNMSE), respectively. TPMSE and TNMSE are defined as
\begin{align}
TPMSE = \textstyle \frac{1}{\widehat{N}_+}\sum_{i:\widehat{\vec{y}}_i>0}{\left|\widehat{\vec{y}}_i-\mu\left(\widehat{\mathcal{X}}_i\right)\right|^2} \\
TNMSE = \textstyle \frac{1}{\widehat{N}_-}\sum_{i:\widehat{\vec{y}}_i<0}{\left|\widehat{\vec{y}}_i-\mu\left(\widehat{\mathcal{X}}_i\right)\right|^2}
\end{align}
where $\widehat{N}_+$ and $\widehat{N}_-$ denote the sizes of the subsets for TPMSE and TNMSE, respectively.

Table \ref{table:gp7DOF} shows the performance of the models in terms of MSE and query time of each of the methods. Note that the average distance to collision is 1.7. We can see the MSEs when using the GP method with the FK kernel are significantly lower (i.e., closer to the GJK/EPA ground truth values) than the other two approximation methods. The FK kernel gives a large boost in modeling accuracy compared to the Gaussian kernel, as was expected \cite{Das2020a}, due to the FK kernel's strong relation with locations in the workspace. The KR method performs worse than GP (Gaussian) in terms of MSE, probably because all points in $\mathcal{X}$ have equal importance with the KR method while GPs assign weights to each point. Fig. \ref{fig:errorHist} shows histograms of absolute errors in estimation of $d_{\mathcal{R}}(\cdot)$ for GP (Gaussian), GP (FK), and KR. We can see that GP (FK) has a larger proportion of lower errors compared to the other methods. 

\begin{table}[!t]
\centering
\begin{tabular*}{0.95\columnwidth}{lllll}
\toprule
Method & MSE & TPMSE & TNMSE & Query Time ($\mu$s) \\
\midrule
GJK/EPA & --- & --- & --- & 466.9 \\
KR & 0.78 & 0.71 & 2.76 & 10.4 \\
GP (Gaussian) & 0.50 & 0.41 & 2.92 & \textbf{1.8} \\
GP (FK) & \textbf{0.06} & \textbf{0.05} & \textbf{0.12} & 6.7 \\
\bottomrule
\end{tabular*}
\caption{Performance in terms of MSE and query time for noise-free GJK/EPA, KR, GP (Gaussian), and GP (FK).}
\label{table:gp7DOF}
\end{table}

\begin{figure*}[!b]
\centering
\subfloat[GJK/EPA]{
	\includegraphics[width=0.18\linewidth, trim={0.8cm 1cm 0.6cm .7cm}, clip]{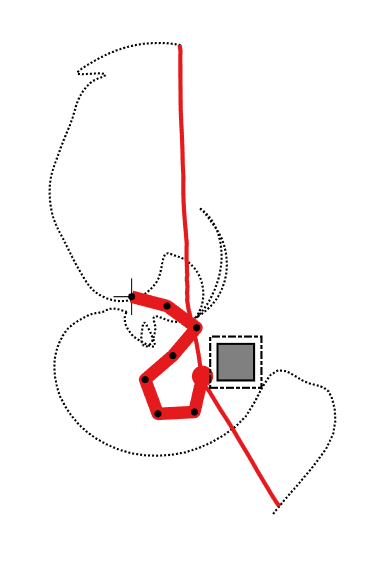}
	}
\hfill
\subfloat[GJK/EPA (Noisy)]{
	\includegraphics[width=0.18\linewidth, trim={0.8cm 1cm 0.6cm .7cm}, clip]{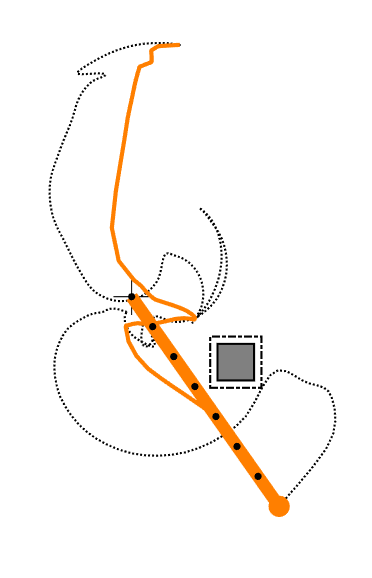}
}
\hfill
\subfloat[KR]{
	\includegraphics[width=0.18\linewidth, trim={0.8cm 1cm 0.6cm .7cm}, clip]{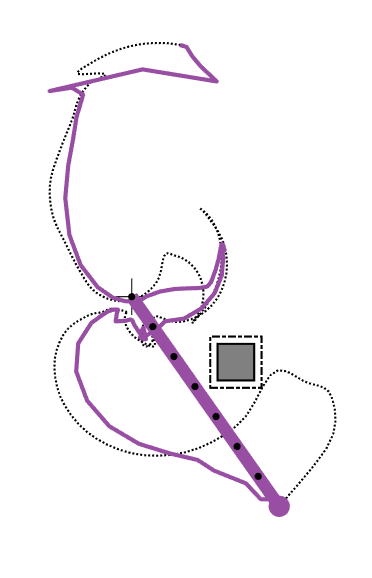}
}
\hfill
\subfloat[GP (Gaussian)]{
	\includegraphics[width=0.18\linewidth, trim={0.8cm 1cm 0.6cm .7cm}, clip]{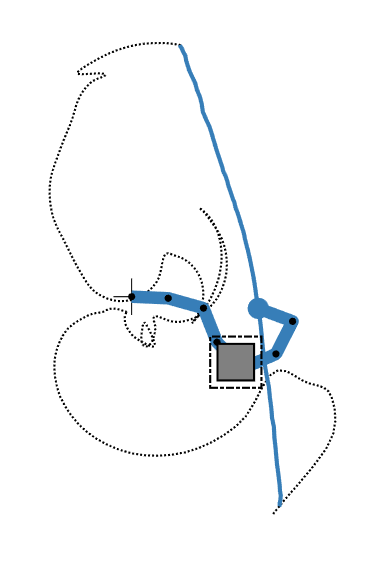}
	}
\hfill
\subfloat[GP (FK)]{
	\includegraphics[width=0.18\linewidth, trim={0.8cm 1cm 0.6cm .7cm}, clip]{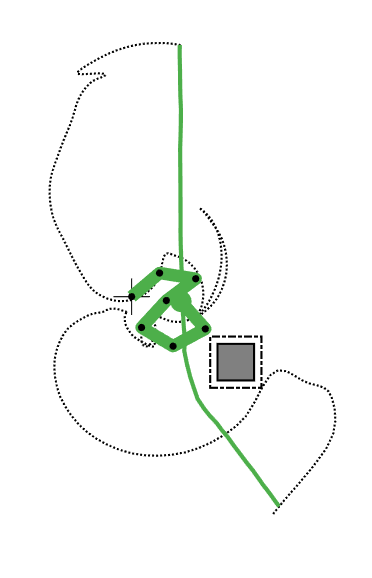}
	}
\caption[Examples of end effector paths before and after optimization when using GJK, GP with Gaussian kernel, GP with FK kernel, and KR for distance estimation.]{Examples of end effector paths before (black curves) and after (colored curves) optimization of Eq. \ref{eq:trajOptimization} (clearance constraint of 0.2) when using (a) GJK/EPA, (b) GJK/EPA with noise, (c) KR, (d) GP with Gaussian kernel, (e) GP with FK kernel for distance estimation. The configuration along the path with the worst distance to collision and the amount of clearance required from the obstacle are also shown.}
\label{fig:gpPaths}
\end{figure*}

GP (FK) is about 3.7 times slower to query than GP (Gaussian) because the FK kernel involves an extra step of evaluating the forward kinematics of the robot and thus takes longer to evaluate than the Gaussian kernel. Also, while the KR method does not require any training time, its query times are longer because the entire training set is used during querying while only a subset of the training set needed for the GP methods. The GP and KR methods used datasets of size $N = 500$, which took approximately 229 ms to create. The training time for the GP (Gaussian) method is about 2.2 seconds, which is approximately 1 second faster than when using the FK kernel.

While GP (FK) is not as fast for training and querying compared to the GP (Gaussian), its MSE is significantly lower. GP (FK) has better modeling performance most likely because the FK kernel has a shape that more appropriately fits the structure of the data, as can be seen by comparing Fig. \ref{fig:fkKernel} to Fig. \ref{fig:distCspace}. Due to the greater accuracy of the GP (FK) model over GP (Gaussian) and KR and the fact that the query time of GP (FK) is still 70 times faster than the ground truth method, GP (FK) is a strong candidate for accurately and efficiently modeling $d_{\mathcal{R}}(\cdot)$.

\subsection{Path Optimization with Clearance Constraint}
\label{sec:gpOptimizationPerformance}
In this section, we utilize the distance estimation techniques described above in a path optimization application. One purpose of path optimization is to minimize a cost, such as path length, while satisfying constraints, such as keeping a certain distance from obstacles, given a motion plan generated by some other method \cite{Schulman2013}. In our experiment, we structure our optimization problem similar to the trajectory optimization formulation in TrajOpt \cite{Schulman2013}. More specifically, the cost is defined as a sum of distances between waypoints under the inequality constraint of distance to collision. 

In our implementation, we minimize the manipulator's end effector trajectory rather than the C-space trajectory in order to avoid trivial solutions such as rotating the manipulator the other way around an obstacle. In this section, we represent the Cartesian position of the end effector in configuration $\vec{x}$ as $FK_{EE}(\vec{x})$. Our optimization problem to find a sequence of $T$ configurations $\Theta=[\theta_1,\ldots,\theta_T]$ where $\theta_t \in \mathbb{R}^D$ is thus 
\begin{equation}
\begin{aligned}
\Theta^* = \argmin_{\Theta} \quad & L\left(\Theta\right)\\
\qquad \text{subject to}\quad & \theta_1 = \theta_{start}, \\
\quad & \theta_T = \theta_{goal}, \\
\quad & \|\theta_{t+1} - \theta_t\| \le \Delta\theta\ \forall t \in [1, \ldots, T-1], \\
\quad &  \min_{t\in[1,\ldots,T]}{d_{\mathcal{R}}(\theta_{t})} \ge d_{min}
\end{aligned} \label{eq:trajOptimization}
\end{equation}
where $\Theta^*$ is the optimized motion plan, $L\left(\Theta\right)=\sum_{t=1}^{T-1}\|FK_{EE}(\theta_{t+1}) - FK_{EE}(\theta_t)\|$ is the distance traveled by the end effector when executing motion plan $\Theta$, $\theta_{start}$ and $\theta_{goal}$ are desired start and goal configurations, $\Delta\theta$ prevents large joint changes, and $d_{min}$ is a desired amount of clearance to maintain from a workspace obstacle. $\dR(\theta_t)$ in the constraint is replaced by each model's estimate.

\begin{table}[!b]
\footnotesize 
\centering
\begin{tabular}{llll}
\toprule
Method & Optim. Time (s) & $L(\Theta)$ & $\min d_{\mathcal{R}}(\Theta) - d_{min}$ \\
\midrule
GJK/EPA & 864.1 & 14.4 & 0.16 \\
GJK/EPA (Noisy) & 240.4 & 16.5 & 0.23 \\
KR & \textbf{29.1} & 26.9 & \textbf{0.29} \\
GP (Gaussian) & 49.2 & 16.1 & -0.12 \\
GP (FK) & 106.8 & \textbf{13.7} & 0.06 \\
\bottomrule
\end{tabular}
\caption{Average optimization time, end effector path length, and minimum satisfaction of clearance constraint after optimization of Eq. \ref{eq:trajOptimization} (clearance constraint) when using GJK/EPA, GJK/EPA with noise, GP (Gaussian kernel), GP (FK kernel), and KR to estimate $d_{\mathcal{R}}(\cdot)$.}
\label{table:gpOpt7DOF}
\end{table}


The constrained optimization problem solver $\mathrm{fmincon}()$ in MATLAB is used to solve Eq. \ref{eq:trajOptimization}. $\theta_{start}$ and $\theta_{goal}$ are randomized on either side of the obstacle to require obstacle avoidance in the motion plan. The initial motion plan is acquired using a Rapidly-Exploring Random Tree motion planner \cite{LaValle2008}, which is then used as the initial seed for the optimization. $d_{min}$ is set to be 0.2 in these experiments.

Table \ref{table:gpOpt7DOF} shows optimization times, optimized path lengths, and the worst satisfaction of the clearance constraint $\min d_{\mathcal{R}}(\Theta)-d_{min}$. Note that $\min d_{\mathcal{R}}(\Theta)-d_{min}$ is positive if the constraint is satisfied. While optimization with GP (Gaussian) and KR is faster than when optimizing with GP (FK), the optimized result is often either in collision or has a long end effector path length. All learning-based methods are significantly faster than GJK/EPA, but using the GP (FK) provides the safest results.

Fig. \ref{fig:gpPaths} shows various examples of the optimized paths using the various methods to approximate $d_{\mathcal{R}}(\cdot)$ in Eq. \ref{eq:trajOptimization}, including a case where the noisy distance measurements are used directly. The original unoptimized end effector path is shown as the dotted curve while the optimized paths are the thicker, colored curves. The configuration along each optimized path with the lowest distance to collision (according to GJK) is displayed, and the amount of desired clearance is represented as the dotted box around the workspace obstacle. It is evident that the optimized trajectories when using the GP (FK) is fairly similar to those when using GJK/EPA, while the other methods yield less optimal paths such as ones with collisions or with longer path lengths.

\begin{figure*}[t]
\centering
\subfloat[GJK/EPA]{
	{\includegraphics[width=0.18\linewidth, trim={0.7cm 2.2cm 0.4cm 1.9cm}, clip]{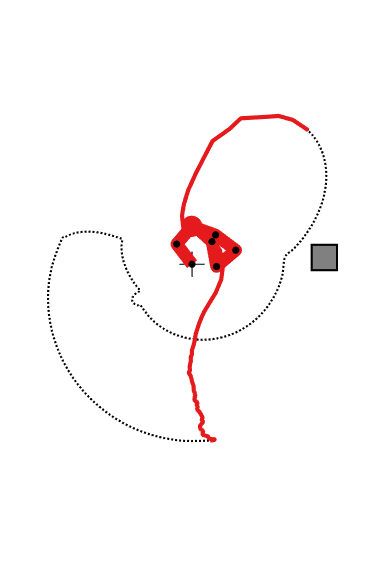}}
	}
\hfill
\subfloat[GJK/EPA (Noisy)]{
	\includegraphics[width=0.18\linewidth, trim={0.7cm 2.2cm 0.4cm 1.9cm}, clip]{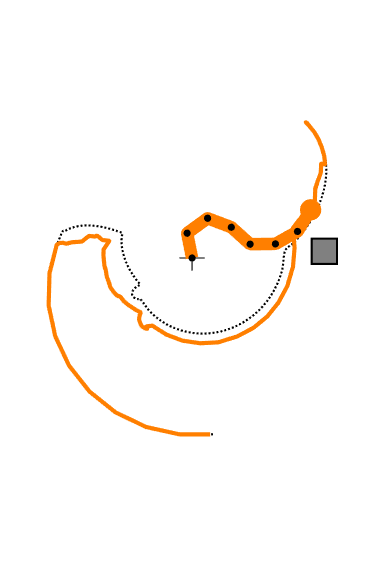}
}
\hfill
\subfloat[KR]{
	\includegraphics[width=0.18\linewidth, trim={0.7cm 2.2cm 0.4cm 1.9cm}, clip]{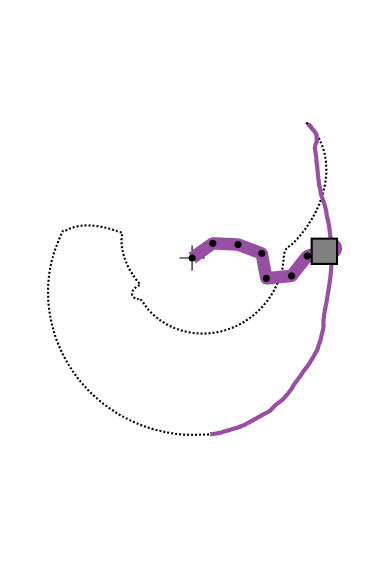}
}
\hfill
\subfloat[GP (Gaussian)]{
	\includegraphics[width=0.18\linewidth, trim={0.7cm 2.2cm 0.4cm 1.9cm}, clip]{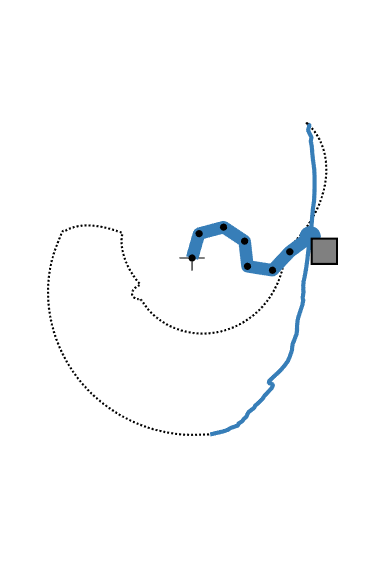}
	}
\hfill
\subfloat[GP (FK)]{
	\includegraphics[width=0.18\linewidth, trim={0.7cm 2.2cm 0.4cm 1.9cm}, clip]{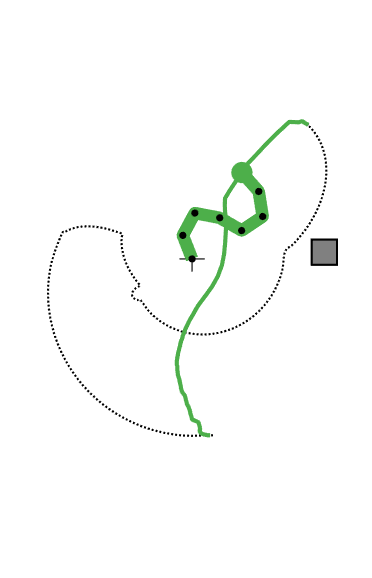}
	}
\caption[Examples of end effector paths before and after optimization when using GJK, GP with Gaussian kernel, GP with FK kernel, and KR for distance estimation.]{Examples of end effector paths before (black curves) and after (colored curves) optimization of Eq. \ref{eq:trajOptimization2} (clearance maximization) when using (a) GJK/EPA, (b) GJK/EPA with noise, (c) KR, (d) GP with Gaussian kernel, (e) GP with FK kernel for distance estimation. The configuration along the path with the worst distance to collision is also shown.}
\label{fig:gpPaths2}
\end{figure*}

\begin{figure*}[!b]
	\centering
	\subfloat[Optimized Path\label{fig:hybridModel}]{%
		{\includegraphics[width=0.48\linewidth, trim={0.6cm 0.6cm 0cm 0.1cm}, clip]{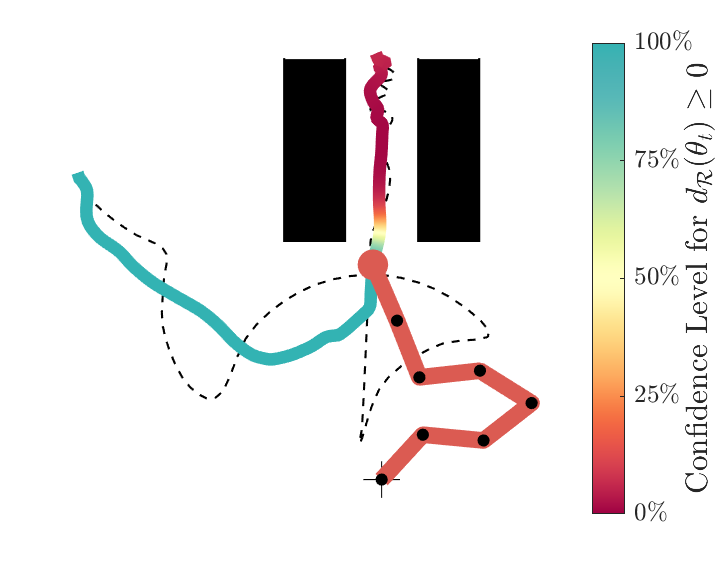}}}
    \hfill
    	\subfloat[Distances\label{fig:hybridDistances}]{%
    {\includegraphics[width=0.48\linewidth, trim={0cm 0.1cm 1cm 0.6cm}, clip]{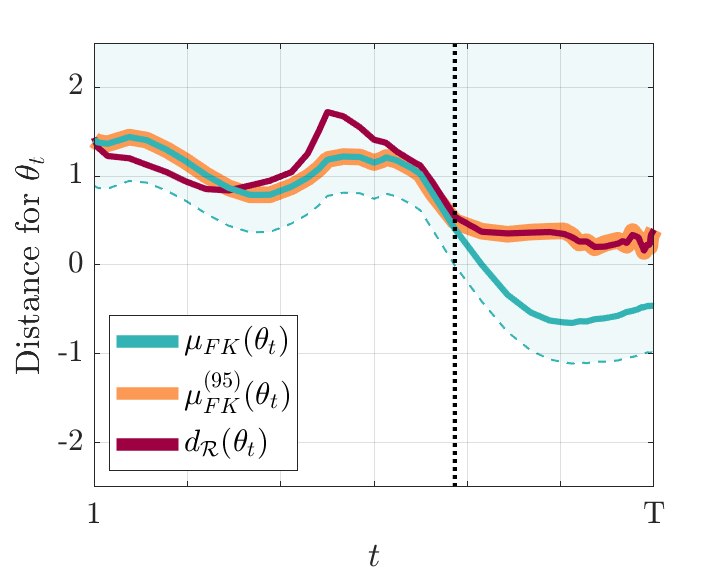}}}
    \caption{(a) The optimized path determined using the hybrid model $\mu^{(95)}_{FK}(\cdot)$ is shown, where the GP's confidence level that the true distance to collision is greater than 0 is represented by the color of the optimized path. The plotted robot configuration marks the point where the hybrid model switches from the GP-based predictions to the sensor-based method. (b) The distances based on the GP model with the FK kernel ($\mu_{FK}(\cdot)$), the hybrid model ($\mu^{(95)}_{FK}(\cdot)$), and the noise-free geometric method $\dR(\cdot)$ for the trajectory $\Theta$ are shown. The shaded region represents the lower-bounded CI for the GP model, and the vertical line shows where the hybrid model switches from GP-based to sensor-based predictions.}
    \label{fig:hybrid}
\end{figure*}

\subsection{Path Optimization with Clearance Maximization}
In this section, we once again apply the distance estimation techniques to path optimization. However, rather than utilizing the distance to collision as a constraint, we include distance in the objective function to increase a trajectory's clearance. We weight the path length objective function used in Eq. \ref{eq:trajOptimization} with an exponentiated distance, which changes the optimization problem to
\begin{equation}
\begin{aligned}
\Theta^* = \argmin_{\Theta} \quad & C\left(\Theta\right)L\left(\Theta\right)\\
\qquad \text{subject to}\quad & \theta_1 = \theta_{start}, \\
\quad & \theta_T = \theta_{goal}, \\
\quad & \|\theta_{t+1} - \theta_t\| \le \Delta\theta\ \forall t \in [1, \ldots, T-1]
\end{aligned} \label{eq:trajOptimization2}
\end{equation}
where $C\left(\Theta\right) = \exp(-\min\dR(\Theta))$. $\dR(\Theta)$ in $C(\Theta)$ is replaced by each distance method's estimate. The exponentiation makes $C(\Theta)$ large for trajectories causing egregious penetration into an obstacle and small for trajectories with larger clearances.

\begin{table}[t]
\footnotesize 
\centering
\begin{tabular}{llll}
\toprule
Method & Optim. Time (s) & $L(\Theta)$ & $\min d_{\mathcal{R}}(\Theta)$ \\
\midrule
GJK/EPA & 481.9 & \textbf{4.3} & \textbf{1.73} \\
GJK/EPA (Noisy) & 56.4 & 32.3 & 0.47 \\
KR & 28.7 & 9.3 & 0.03 \\
GP (Gaussian) & \textbf{15.7} & 4.5 & -0.25 \\
GP (FK) & 51.0 & 4.9 & 1.24 \\
\bottomrule
\end{tabular}
\caption[Results for optimal paths using GJK \cite{Gilbert1988,VandenBergen2001}, GP (Gaussian kernel) \cite{Rasmussen2006}, GP (FK kernel) \cite{Rasmussen2006, Das2020a}, KR \cite{Nadaraya1964,Watson1964} to calculate $d_{\mathcal{R}}(\cdot)$ are presented. We include the average times to achieve the optimized plan, the average final cost, and the average worst distance to collision in each optimized path.]{Average optimization time, end effector path length, and minimum distance to collision after optimization of Eq. \ref{eq:trajOptimization2} (clearance maximization) when using GJK/EPA, GJK/EPA with noise, GP (Gaussian kernel), GP (FK kernel), and KR to estimate $d_{\mathcal{R}}(\cdot)$.}
\label{table:gpOpt7DOF2}
\end{table}

Table \ref{table:gpOpt7DOF2} shows average optimization times, path lengths, and minimum distances to collision for each method. GP (FK) has a higher amount of clearance than GP (Gaussian), KR, and noisy GJK/EPA methods, and in fact achieves results closer to the noise-free geometric method both in terms of clearance and path length despite being trained on noisy measurements. The noisy version of GJK/EPA yielded suboptimal solutions with large end effector path lengths and smaller clearances than GP (FK). GP (Gaussian) and KR have shorter optimization times because the optimization solver often quickly converges to an invalid solution (as can be seen from the lower minimum distances to collision). The paths in Fig. \ref{fig:gpPaths2} exemplify these results.

\subsection{Confidence-Based Hybrid Model}

We apply our proposed hybrid model to a scenario involving a narrow passage, where the manipulator must reach between two nearby obstacles. Inaccuracy in the distance estimation model may inadvertently lead to collision between these obstacles, so the hybrid model's predictions can potentially handle this scenario by switching to the geometric method if the GP model is not confident.

Fig. \ref{fig:hybridModel} shows an example optimized trajectory achieved with the hybrid model $\mu^{(95)}_{FK}(\cdot)$. The plotted robot configuration denotes the transition point where the hybrid model switches from GP-based predictions to noisy, sensor-based distance evaluations $\tilde{d}_\mathcal{R}(\cdot)$. We used $N_{\tilde{d}}=5$ evaluations of $\tilde{d}_\mathcal{R}(\cdot)$ per configuration when the hybrid model did not use a GP-based prediction. The colors of the path denote the confidence level of the GP that the true distance to collision is greater than 0, which we determine via MATLAB's $\mathrm{normcdf}()$. We can observe that the GP model is fairly confident when far from the obstacles, but the confidence level rapidly drops when the manipulator needs to move into the narrow passage.

Fig. \ref{fig:hybridDistances} shows the GP-based distance estimates $\mu_{FK}(\Theta)$, the hybrid estimates $\mu^{(95)}_{FK}(\Theta)$, and the true distance to collision $\dR(\Theta)$. The shaded region represents the lower-bounded CI for the GP model. We can see that $\mu_{FK}(\Theta)$ deviates from $\dR(\Theta)$ as the distance decreases, but the hybrid model $\mu^{(95)}_{FK}(\Theta)$ manages to produce accurate measurements of distance by switching to the noisy distance measurements when the GP loses confidence.


\section{Conclusion}
In this paper, we propose a method to create a model for evaluating a robot manipulator's distance to collision. Our approach utilizes GP regression and the FK kernel, a similarity function designed to compare configurations of a robot manipulator. Noisy distance measurements are used to train the models.

Querying the GP model with the FK kernel obtained up to 70 times faster distance estimates than the geometric method. The accuracy of this GP model significantly surpassed that of other regression techniques with up to 13 times lower MSE, even when trained on noisy distance values.  The GP model with the FK kernel achieved optimized paths 9 times faster but of similar quality as was achieved by a noise-free geometric method.

We also introduce a confidence-based hybrid model that uses the GP mean as a predicted distance in areas of high confidence and an averaging method over sensor-based distances in areas of lower confidence. We demonstrate the capability of this technique in a narrow passage environment and realize the hybrid model switches to the sensor-based method as it gets closer to the obstacles.

Using GP regression with the FK kernel is thus a promising approach for creating a model for distance evaluation, even when working with noisy measurements. More efficient sampling, online training and updating, and parallelization are left for future work.

\bibliographystyle{IEEEtran}
\bibliography{references}

\begin{thebibliography}{10}
\providecommand{\url}[1]{#1}
\csname url@samestyle\endcsname
\providecommand{\newblock}{\relax}
\providecommand{\bibinfo}[2]{#2}
\providecommand{\BIBentrySTDinterwordspacing}{\spaceskip=0pt\relax}
\providecommand{\BIBentryALTinterwordstretchfactor}{4}
\providecommand{\BIBentryALTinterwordspacing}{\spaceskip=\fontdimen2\font plus
\BIBentryALTinterwordstretchfactor\fontdimen3\font minus
  \fontdimen4\font\relax}
\providecommand{\BIBforeignlanguage}[2]{{%
\expandafter\ifx\csname l@#1\endcsname\relax
\typeout{** WARNING: IEEEtran.bst: No hyphenation pattern has been}%
\typeout{** loaded for the language `#1'. Using the pattern for}%
\typeout{** the default language instead.}%
\else
\language=\csname l@#1\endcsname
\fi
#2}}
\providecommand{\BIBdecl}{\relax}
\BIBdecl

\bibitem{Das2020}
N.~Das and M.~Yip, ``{Learning-Based Proxy Collision Detection for Robot Motion
  Planning Applications},'' \emph{IEEE Transactions on Robotics}, pp. 1--19,
  2020.

\bibitem{Flacco2012}
F.~Flacco and A.~D. Luca, ``{A Depth Space Approach to Human-Robot Collision
  Avoidance},'' in \emph{IEEE International Conference on Robotics and
  Automation}, 2012.

\bibitem{Schiavi2009}
R.~Schiavi, F.~Flacco, and A.~Bicchi, ``{Integration of Active and Passive
  Compliance Control for Safe Human-Robot Coexistence},'' in \emph{IEEE
  International Conference on Robotics and Automation}, 2009.

\bibitem{Pan2013}
J.~Pan, X.~Zhang, and D.~Manocha, ``{Efficient Penetration Depth Approximation
  using Active Learning},'' \emph{ACM Transactions on Graphics}, vol.~32,
  no.~6, 2013.

\bibitem{Lin2003}
M.~C. Lin and D.~Manocha, ``{Collision and Proximity Queries},'' in
  \emph{Handbook of Discrete and Computational Geometry}, 3rd~ed.\hskip 1em
  plus 0.5em minus 0.4em\relax CRC Press, 2003, pp. 1029--1056.

\bibitem{Han2019}
\BIBentryALTinterwordspacing
Y.~Han, W.~Zhao, J.~Pan, Z.~Ye, R.~Yi, and Y.-j. Liu, ``{A Configuration-Space
  Decomposition Scheme for Learning-based Collision Checking},'' \emph{arXiv
  preprint arXiv:1911.08581 [cs]}, 2019. [Online]. Available:
  \url{https://arxiv.org/abs/1911.08581}
\BIBentrySTDinterwordspacing

\bibitem{Pan2012}
J.~Pan, S.~Chitta, and D.~Manocha, ``{FCL: A general purpose library for
  collision and proximity queries},'' \emph{Proceedings - IEEE International
  Conference on Robotics and Automation}, pp. 3859--3866, 2012.

\bibitem{Kyriakopoulos1990}
\BIBentryALTinterwordspacing
K.~J. Kyriakopoulos and G.~N. Saridis, ``{Minimum Distance Estimation and
  Prediction Under Uncertainty for Online Robotic Motion Planning},''
  \emph{IFAC Proceedings Volumes}, vol.~23, no.~8, pp. 93--98, 1990. [Online].
  Available: \url{http://dx.doi.org/10.1016/S1474-6670(17)51718-8}
\BIBentrySTDinterwordspacing

\bibitem{Safeea2017}
M.~Safeea, N.~Mendes, and P.~Neto, ``{Minimum Distance Calculation for Safe
  Human Robot Interaction},'' \emph{Procedia Manufacturing}, vol.~11, no. June,
  pp. 99--106, 2017.

\bibitem{Gilbert1988}
E.~G. Gilbert, D.~W. Johnson, and S.~S. Keerthi, ``{A Fast Procedure for
  Computing the Distance Between Complex Objects in Three-Dimensional Space},''
  \emph{IEEE Journal on Robotics and Automation}, vol.~4, no.~2, pp. 193--203,
  1988.

\bibitem{Rakita2018}
D.~Rakita, B.~Mutlu, and M.~Gleicher, ``{RelaxedIK: Real-time Synthesis of
  Accurate and Feasible Robot Arm Motion},'' in \emph{Robotics: Science and
  Systems}, 2018.

\bibitem{Kew2019}
J.~C. Kew, B.~Ichter, and T.-w.~E. Lee, ``{Neural Collision Clearance Estimator
  for Fast Robot Motion Planning},'' \emph{arXiv preprint arXiv:1910.05917
  [cs.RO]}, 2019.

\bibitem{Hofmann2008}
T.~Hofmann, B.~Sch{\"{o}}lkopf, and A.~J. Smola, ``{Kernel methods in machine
  learning},'' \emph{Annals of Statistics}, vol.~36, no.~3, pp. 1171--1220,
  2008.

\bibitem{Altman1992}
N.~S. Altman, ``{An introduction to kernel and nearest-neighbor nonparametric
  regression},'' \emph{American Statistician}, vol.~46, no.~3, pp. 175--185,
  1992.

\bibitem{Minh}
H.~Q. Minh, P.~Niyogi, and Y.~Yao, ``{Mercer's Theorem, Feature Maps, and
  Smoothing},'' in \emph{International Conference on Computational Learning
  Theory}, 2006, pp. 154--168.

\bibitem{Rasmussen2006}
C.~E. Rasmussen and C.~K.~I. Williams, \emph{{Gaussian Processes for Machine
  Learning}}.\hskip 1em plus 0.5em minus 0.4em\relax The MIT Press, 2006.

\bibitem{BelancheMunoz2013}
L.~A. {Belanche Munoz}, ``{Developments in Kernel Design},'' in \emph{European
  Symposium on Artificial Neural Networks, Computational Intelligence and
  Machine Learning}, 2013.

\bibitem{Smola1998}
A.~J. Smola, B.~Scholkopf, and K.~R. Muller, ``{The connection between
  regularization operators and support vector kernels},'' \emph{Neural
  Networks}, vol.~11, no.~4, pp. 637--649, 1998.

\bibitem{Das2020a}
N.~Das and M.~C. Yip, ``{Forward Kinematics Kernel for Improved Proxy Collision
  Checking},'' \emph{IEEE Robotics and Automation Letters}, vol.~5, no.~2, pp.
  2349--2356, 2020.

\bibitem{Das2017}
\BIBentryALTinterwordspacing
N.~Das, N.~Gupta, and M.~Yip, ``{Fastron: An Online Learning-Based Model and
  Active Learning Strategy for Proxy Collision Detection},'' \emph{Proceedings
  of the 1st Annual Conference on Robot Learning}, vol.~78, pp. 496--504, 2017.
  [Online]. Available: \url{http://proceedings.mlr.press/v78/das17a.html}
\BIBentrySTDinterwordspacing

\bibitem{Fahrmeir2013}
L.~Fahrmeir, T.~Kneib, S.~Lang, and B.~Marx, \emph{{Regression Models, Methods
  and Applications}}.\hskip 1em plus 0.5em minus 0.4em\relax Springer, 2013.

\bibitem{Nadaraya1964}
E.~Nadaraya, ``{On Estimating Regression},'' \emph{Theory of Probability and
  Its Applications}, vol.~9, no.~1, pp. 157--159, 1964.

\bibitem{Watson1964}
G.~S. Watson, ``{Smooth Regression Analysis},'' \emph{Sankhya: The Indian
  Journal of Statistics}, vol.~26, no.~4, pp. 359--372, 1964.

\bibitem{Altman1988}
D.~G. Altman and M.~J. Gardner, ``{Calculating confidence intervals for
  regression},'' \emph{British Medical Journal}, vol. 296, no. April, pp.
  1238--1242, 1988.

\bibitem{Wilcox2020}
B.~Wilcox and M.~C. Yip, ``{SOLAR-GP: Sparse Online Locally Adaptive Regression
  Using Gaussian Processes for Bayesian Robot Model Learning and Control},''
  \emph{IEEE Robotics and Automation Letters}, vol.~5, no.~2, pp. 2832--2839,
  2020.

\bibitem{VandenBergen2001}
\BIBentryALTinterwordspacing
G.~van~den Bergen, ``{Proximity queries and penetration depth computation on 3d
  game objects},'' \emph{Game developers conference}, 2001. [Online].
  Available:
  \url{https://graphics.stanford.edu/courses/cs468-01-fall/Papers/van-den-bergen.pdf}
\BIBentrySTDinterwordspacing

\bibitem{Schulman2013}
J.~Schulman, J.~Ho, A.~Lee, I.~Awwal, H.~Bradlow, and P.~Abbeel, ``{Finding
  Locally Optimal, Collision-Free Trajectories with Sequential Convex
  Optimization},'' in \emph{Robotics: Science and Systems}, 2013.

\bibitem{LaValle2008}
S.~M. LaValle, ``{Rapidly-Exploring Random Trees: A New Tool for Path
  Planning},'' \emph{Animal Genetics}, vol.~39, no.~5, pp. 561--563, 2008.

\end{thebibliography}
\end{document}